# Continuous learning of face attribute synthesis

Xin Ning, Weijun Li, Xiaoli Dong
Institute of Semiconductors, Chinese Academy of Sciences
Cognitive Computing Technology Joint Laboratory, Wave Group
Beijing, China
{ninxing, liweijun, dongxiaoli}@semi.ac.cn

Shaohui Xu, Fangzhe Nan
Cognitive Computing Technology Joint Laboratory, Wave Group
Beijing, China
{xushaohui, nanfangzhe}@wavewisdom-bj.com

Yuanzhou Yao
College of Information Engineering, Sichuan Agricultural University
Ya'an, Sichuan, China
yaoyuanzhou@stu.sicau.edu.cn

*Abstract*—The generative adversarial network (GAN) exhibits great superiority in the face attribute synthesis task. However, existing methods have very limited effects on the expansion of new attributes. To overcome the limitations of a single network in new attribute synthesis, a continuous learning method for face attribute synthesis is proposed in this work. First, the feature vector of the input image is extracted and attribute direction regression is performed in the feature space to obtain the axes of different attributes. The feature vector is then linearly guided along the axis so that images with target attributes can be synthesized by the decoder. Finally, to make the network capable of continuous learning, the orthogonal direction modification module is used to extend the newly-added attributes. Experimental results show that the proposed method can endow a single network with the ability to learn attributes continuously, and, as compared to those produced by the current state-of-the-art methods, the synthetic attributes have higher accuracy.

## I. INTRODUCTION

In recent years, with the continuous development of deep neural networks, related research in image field has made great progress. However, in most neural networks, the output largely depends on the input, and the models have fixed and consistent mapping rules [1]. Unlike the cognitive perception ability of human beings and primates, once the training of a deep neural network is completed, the parameters will be fixed. When a new dataset is used to train an existing model, it will lose the ability to distinguish the original dataset, which is called "catastrophic forgetting" [2]. Therefore, determining to make neural networks retain the ability to distinguish old data when learning new mapping rules has recently become popular research topic.

For existing attribute synthesis tasks, researchers usually modify the feature representation of an input image to generate images with different attribute expressions, which plays an active role in task research in a small sample environment. Since the generative adversarial network (GAN) [3] was first proposed, it has attracted widespread attention and has become the current mainstream method of attribute synthesis. Methods that use GAN to synthesize face image attributes can be divided into two types, namely style mixing models and image translation models. Style mixing models extract feature vector of an input image, linearly interpolates the feature to obtain the middle state of the input attribute, and then use the generator to reconstruct an image with the gradual change of target attributes. For example, PGGAN [4] increases the resolution of generated images by gradually and synchronously increasing the network layer, and the feature distribution is more centralized via a normalization operation, and a balanced learning rate is used to ensure the dynamic updating and consistency of the weight learning rate. Based on PG-GAN, StyleGAN [5] further improves the quality of the generated image by adding a mapping network and encoding an input vector as an intermediate vector, performing an AdaIN operation after upsampling and convolution to fully decouple different attributes, and adding scaled noise to each channel to make the generated image more realistic and diverse. StyleGAN2 [6] features a redesigned normalization method of generator that, eliminates artifacts that appear in images generated by StyleGAN. In addition, StyleGAN2 can also project an input image to latent space. Although the style mixing models can freely control the changing process of target attributes via the linear interpolation of features, the non-target attributes are also changed during feature interpolation, resulting in inaccurate attribute operations.

Most image translation models have an encoding-decoding architecture; by adding supervision information to the target domain during the decoding process, the synthesized image will have target attributes. For example, in 2014, Mirza proposed CGAN [7], which is an improvement of GAN that adds condition information to the network input. In the CGAN generator, the input noise Z and supervision information y are connected through a fully connected layer as the input of a hidden layer. Similarly, the supervision information y and the generated image are connected together as an input of a discriminator network. By adding additional information to the original generator and discriminator, the GAN network can use images and corresponding labels for training, and uses the given label information to generate a specific image in a test phase. Based on CGAN, Phillip [8] proposed the pix2pix model that use paired data for image translation. The supervision information y is replaced with image x given by a user, which is then used as the input of the generator to obtain the generated image G (x). G (x) and x are then merged together based on the channel dimension, and the predicted probability value is calculated from the discriminator. In addition, the real image and y are also merged based on the channel dimension and predicted by the discriminator. Choi [9] proposed StarGAN to achieve the transfer to multiple fields; similar to that of CGAN, the discriminator of StarGAN not only must learn to discriminate whether the sample is real, but must also judge which domain the images is. Lin [10] proposed STGAN, which uses selective transfer units (STUs) after each encoder and the difference vector of the target attribute and source image as input, target image is reconstructed by the decoder. Via the combination of STUs and an encoder-decoder structure to adaptively select and modify encoder features, STGAN improves the quality of generated images while also improving the accuracy of attribute editing. Although research on image translation models has made great progress, GAN can only fit the fixed mapping relationship between an input image and target attributes, and the existing mapping rules have a very limited effect on the expansion of new attributes.

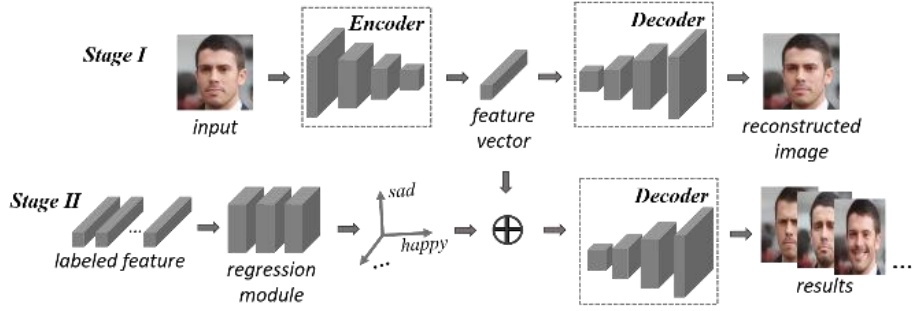

Fig. 1. Network structure of attribute composition.

In summation, regardless of the use of an image translation model or style mixing model, single models have no expansion capabilities. Once parameters are determined, the ability to manipulate newly added attributes cannot be achieved, and the attributes of the synthesized image have limited accuracy. Thus, to overcome this problem, a continuous learning attribute generation model is proposed in the present work. The change directions of different attributes in the feature space are explored, and orthogonal weight modification operations are used to obtain more accurate manipulation results and achieve the expansion ability of new attributes.

## II. THE PROPOSED METHOD

### A. Overview

As illustrated in Figure.1, the continuous synthesis of attributes can be divided into two stages, namely feature extraction and attribute axis regression. In the first stage, the encoder was trained, so that the features of the input image can be extracted. In the first stage, the trained decoder is used as a part of the network to assist the training of the encoder, so that the encoder can extract more comprehensive features from the input image. In the second stage, the features of face data with attribute labels are extracted from the encoder in the first stage, and the generalized linear model is used to fit the directions of different attributes. In order to synthesize the target attribute image, linearly guide the characteristics of the input image and use the decoder to generate the target images.

### B. Feature Extraction

Image representation can be decomposed into content information and style information [10]. The codec network structure for network training is used in this work, and attribute regression operations is performed on the extracted features to obtain the attribute change axis. Because the network does not initially have any supervision information, the generator $G$ is trained as the decoder to assist in the training of the encoding network. Constraint $min\ ||f(x^r) - f(G(E(x^r)))||$ is used to ensure that the input and generated images are as consistent as possible, where $x^r$ represents the input image, $E$ is used to extract the feature vector of the input image, and $f$ represents the calculation process of the difference between the input and generated images. The VGG16 network is used as the main structure of the encoder to extract feature vectors from the input images, and in order to make the extracted feature vector contain all the attributes of the input, the pixel loss and perceptual loss between the input image and the reconstructed image are calculated, and used the superimposed result as a loss function to optimize the parameters of the encoder.

### C. Attribute Axis Regression

After training, a one-to-one mapping relationship $vec = E(x^r)$ is established between the feature vector and input. To obtain the change directions of different attributes, a labeled feature dataset $\{vec, y\}$ is used to explore the attribute direction. And then the Ordinary Least Squares (OLS) model is used for the linear regression of the paired data, and different attribute axes are regarded as linear combinations of all dimension vectors, which is expressed as:

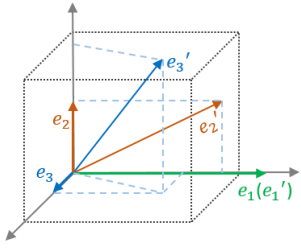

Fig. 2. An orthogonal example in 3D space.

$$f_{vec}(w, x) = w_0 + w_1 x_1 + \cdots + w_p x_p, \quad (1)$$

where $x$ represents the value in each dimension in the attribute direction, and $p$ represents the total number of dimensions in the attribute direction. When $M$ in (2) reaches the minimum value, the resulting attribute direction is considered to be the best combination of the vectors in each dimension.

$$M = \sum_{i=1}^{n}(y - y')^2 = (y - w_0 - \sum_{i=1}^{n} w_i x_i)^2 \quad (2)$$

To obtain more accurate face attribute changes from the regression attribute direction, the operation of orthogonal attribute direction modification is used to fully decouple attributes. First, nonlinear correlations are assumed between different attribute axes, which can avoid the singularity of the coefficient matrix. Additionally, because most attribute directions are intersecting, when an attribute is manipulated, other irrelevant attributes will also change; this is called "feature entanglement." For example, in face editing, if only beard features are added to the face of an input image, the synthetic image will change to be a male; if only the skin quality of the face of an input image is changed, the generated image will be more feminine. To solve the problem of feature entanglement between different attributes, the feature directions of different attributes are dealt with in an orthogonal way, and they are projected to a new direction perpendicular to the other feature directions to achieve the directivity change of a single attribute.

The orthogonality of attributes in 3D space is presented as an example in Figure 2. Suppose that there are three independent linear basic directions, $e_1'$, $e_2'$, and $e_3'$ that respectively represent three basic attributes; $e_1'$ is specified as the basic direction of $e_1$. $e_2'$ is used to subtract its projection on $e_1$, and keeping its component perpendicular to $e_1$ as $e_2$.

$$e_2 = e_2' - \frac{e_1^T e_2'}{e_1^T e_1} e_1 \quad (3)$$

Similarly, using $e_3'$ is used to subtract its projection component on $e_1$ and $e_2$, and keeping its vertical part as $e_3$.

$$e_1 = \frac{e_1}{||e_1||} \quad e_2 = \frac{e_2}{||e_2||} \quad e_3 = \frac{e_3}{||e_3||} \quad (4)$$

Finally, as given by (4), the different attribute directions are normalized to obtain the final direction.

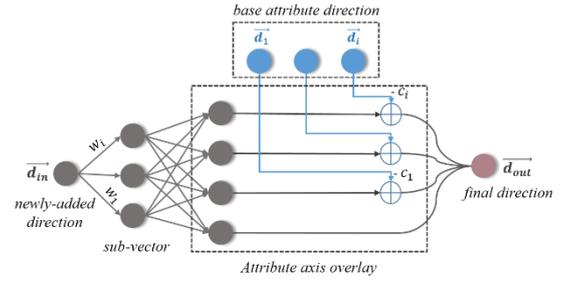

Fig. 3. New property orientation extension.

### D. Attribute continuous learning

For face editing, most generated models do not have the ability of continuous learning; when processing newly added attributes, the parameters of the trained model must be retrained each time, which wastes many human and material resources. To endow a single generation model with the ability to continuously edit different attributes, the trained model is used to expand the newly added attributes. After learning 6 facial expression attributes, 4 other attributes, namely "beard," "eyebrow," "mouth," and "eyes," are included. Specifically, after obtaining the direction vector of a basic attribute, the corresponding relationship between the new attribute label and feature vector is established through feature vector extraction, and the OLS operation is used to obtain the new attribute direction $\overrightarrow{d_{in}}$. Firstly, the newly added attributes are projected in directions perpendicular to each basic attribute. As shown in (5), $\overrightarrow{d_{in}}$ is first decomposed into sub-vectors of different intensities,

$$\overrightarrow{d_{in}} = w_1 * \overrightarrow{d_{in}} + w_2 * \overrightarrow{d_{in}} + \cdots + w_n * \overrightarrow{d_{in}}, \quad (5)$$

where $\sum_{i=1}^{n} w_i = 1$, and $n$ represents the number of basic attributes.

Secondly, in order to obtain the final vector of the newly added attribute, each sub-vector is projected in the direction of the basic attribute, and the additional direction will be used as the main direction of the newly added attribute. As presented in Figure.3, sub-vectors are projected onto basis vectors and get a direction perpendicular to each basic vector, and the results of all subcomponents are superposed to obtain the direction vector of the newly-added attribute. The formula is expressed as:

$$\overrightarrow{d_{out}} = \sum_{i=1}^{n+1} p(w_i * \overrightarrow{d_{in}}) + (-c_i * \overrightarrow{d_{in}}), \quad (6)$$

where $p(\bullet)$ represents the projection operation for each sub-component.

Finally, $\overrightarrow{d_{out}}$ is normalized to obtain the final direction of the newly-added attribute.

### E. Loss Function

The VGG16 [12] network is used to calculate the pixel difference and perceptual distance between the input and generated images. The loss function uses both perceptual loss and pixel loss, which can be expressed as:

$$L = L_{pixel} + L_{perceptual}, \quad (7)$$

The definition of perceptual loss is as follows:

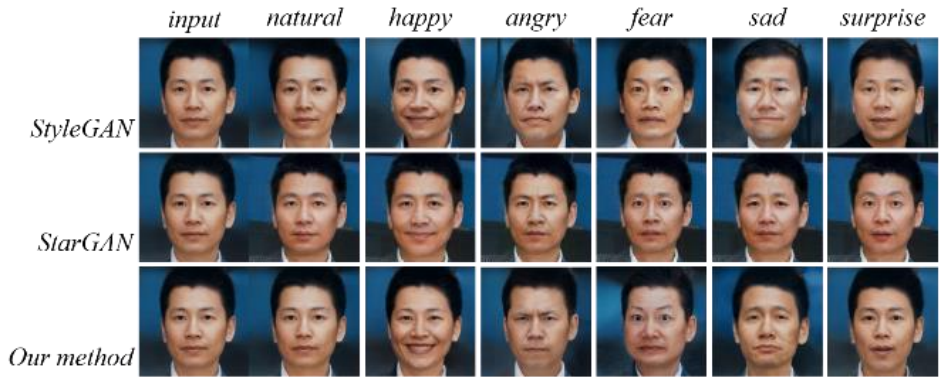

Fig. 4. Comparison of the experimental results of six expressions on the AffectNet database generated by different algorithms.

$$L_{perceptual}(x^r, x^s) = \sum_{i=1}^{4} \frac{1}{C_i H_i W_i} ||E_i(x^r) - E_i(x^s)||, \quad (8)$$

where $E_i(\bullet)$ is the output of the i-th convolutional layer and $C_i H_i W_i$ is the number of output features from the i-th layer. Relative to the L1 loss function, the L2 loss function amplifies the gap between the maximum error and minimum error and is more sensitive. The VGG16 network is used to calculate the perceptual loss, and the L2 distances of the conv1_1, conv1_2, conv3_2, and conv4_2 layers are calculated, so that the generated image and input images are as semantically consistent as possible. Various loss functions are used for testing, such as log-cosh, MAE, MSE, and MS-SSIM loss. Among them, MS-SSIM loss is a multi-scale structural similarity loss function, that not only considers the brightness, contrast, and structural factors of an image, but also the image resolution, whereas MSE loss prefers some similarities for detail calculation; therefore, the MS-SSIM and MSE loss functions are combined as the pixel loss.

### III. EXPERIMENTS

#### A. Dataset and details

To prove the effectiveness of the proposed method, the CelebA [13] and AffectNet [14] datasets were used for experiments. CelebA contains 202,599 face images of 10,177 identities, each of which contains 40-dimensional binary attribute tags that are widely used in the research of face attributes. The AffectNet dataset contains images of more than 1 million people collected from the internet, and the attributes of about half of the images (440,000) regarding 7 facial expressions have been manually marked.

In the experiment, the decoder (generator) used the pre-trained StyleGAN model, and the encoder used the VGG16 network. After face detection and the alignment of the input images, the size was unified to 1024×1024 pixels, and the features were encoded to 18×512 dimensions. The TensorFlow framework and 1080Ti×2 were employed on a Ubuntu 16.04 server for model training. The Adam optimizer was used, β_1 = 0.9, β_2 = 0.999, and epsilon = 1e-08 were set, and the initial learning rate was set to 0.001 and reduced by half every 500 epochs.

#### B. Qualitative results

##### 1) Basic attribute synthesis

31,528 facial expression data including "natural," "happy," "angry," "fear," "sad," and "surprised" expressions were selected from AffectNet to train the facial expression synthesis model. In this experiment, the style transfer model StarGAN and the style mixing model StyleGAN were selected for comparison, and the experimental results are presented in Figure 4. Several methods modified the target attributes. For the algorithm based on StarGAN, the quality of the generated images was not high, it exhibited the insufficient manipulation of expressions such as "happy" and "sad," and local fuzzy conditions occurred; in particular, the mouth areas of the generated images were generally blurred. For the style-mixing model, the resolutions of the images generated by StyleGAN were high; however, attribute manipulation was realized by feature weighting and the manipulation of the target attribute was relatively loose. Thus, the area outside the target attribute was easily changed; for example, the "angry," "sad," and "surprised" expressions were modified, the facial contours changed significantly, and identity differences between the synthesized images and input image were obvious. In contrast, the images generated by the proposed method had relatively high resolutions, and more attention was paid to the manipulation of the target attributes. The generated results conformed to intuitive logic, and the synthesis results of the "natural," "happy," and "angry" expressions were closer to the real biological characteristics.

##### 2) Continuous expansion

To verify the effectiveness of continuous learning, the previously-used 6 expressions were used as basic attributes and the experiment was expanded to include 4 additional attributes. Due to the large differences in the numbers of different attributes in the CelebA dataset, the 4 attributes of "beard," "eyebrow," "mouth," and "eye" were selected, each of which had about 3,000 samples. First, the steps outlined in section 2.1 were followed to establish correspondences between the attribute labels and feature vectors, and least squares regression was then performed on the feature vectors and attribute labels to obtain the newly-added attribute axes. The additional 4 attributes were sequentially expanded by subtracting the projection on the direction of each expression attribute from the newly-added attribute axis to obtain the attribute direction that was not related to the expression, and the experimental results are presented in Figure 5(a). It is evident that the proposed algorithm achieved satisfactory results on the expansion of these 4 attributes, and could smoothly change between different attributes. In addition, the StyleGAN model was also used to expand and synthesize the newly-added attributes, and the experimental results are exhibited in Figure 5(b). While modifying the target attribute, the non-target attribute also changed significantly, and the result of attribute synthesis was largely dependent on the selected auxiliary image.

TABLE I. ACCURACIES OF THE SYNTHETIC ATTRIBUTES OF DIFFERENT ALGORITHMS.

|  | natural | happy | angry | fear | sad | surprise | beard | mouth | eyebrow | eye |
|---|---|---|---|---|---|---|---|---|---|---|
| AttGAN[17] | 0.49 | 0.56 | 0.51 | 0.43 | 0.49 | 0.45 | 0.52 | 0.47 | 0.40 | 0.43 |
| StarGAN[9] | 0.60 | 0.64 | 0.54 | 0.59 | 0.45 | 0.60 | 0.64 | 0.54 | 0.59 | 0.45 |
| StyleGAN[5] | 0.55 | 0.49 | 0.42 | 0.38 | 0.33 | 0.55 | 0.49 | 0.42 | 0.38 | 0.33 |
| **Proposed** | **0.62** | **0.78** | **0.57** | **0.51** | **0.67** | **0.62** | **0.78** | **0.57** | **0.51** | **0.67** |

*C. Quantitative evaluation*

To verify the accuracy of the attribute synthesis of different algorithms, ResNet50 was used to obtain the classification functions of the 10 attributes of "natural," "happy," "angry," "fear," "sad," "surprise," "beards," "mouths," "eyebrows," and "eyes" on the training set, which could achieve 86.3% accuracy on the verification dataset. The first 6 expression attributes were used as basic attributes for training. Next, feature weighting and the algorithm proposed in this paper were used to continuously learn the last 4 attributes. Considering that the StarGAN network is not scalable, the StarGAN generation model was retrained on the "beards," "mouths," "eyebrows," and "eyes" attributes on an additional model. Different algorithms were then used to generate 100 images of each of the 10 attributes, and the classification model was then used for testing.

The experimental results are exhibited in Table 1, it can be seen that the StarGAN-based method has a much higher accuracy than the StyleGAN method in the synthesis of regular expression such as "happy" and "angry", because it requires retraining the model on newly added attribute data. But it cannot be ignored that the quality of the generated image is lower than that of StyleGAN which uses attribute layering and adding convolutional layers to improve the quality of the generated image. In order to flexibly control the process of attribute synthesis and improve the quality of attribute synthesis, the StyleGAN generator is used as a decoder in the proposed method, and utilize orthogonal operation to ensure a higher accuracy of attribute synthesis. Furthermore, the method proposed in this paper is to adjust the direction of the newly added attribute according to the direction of the trained basic attribute, when operating on a newly added attribute, the original attribute is hardly affected.

Overall, it can be summarized from Table.1 that the proposed method achieved the best accuracy of the classification for most attributes, especially for the "happy" attribute for which the accuracy reached greater than 70%.

IV. ABLATION STUDY

In order to verify the quality of images generated using different loss functions, a variety of loss functions were used for experiments. Furthermore, in order to judge the performance of the network on other data sets, 3,257 car images were collected from the Internet, including "left", "right", "front", "black", "white" and "blue" 6 attributes, and select the orientation change as the basic attribute, and the color attribute as the extended attribute.

TABLE II. COMPARISON OF IMAGE QUALITY AFTER RECONSTRUCTION BY DIFFERENT LOSS FUNCTIONS.

| Loss Function | PSNR | SSIM |
|---|---|---|
| Log-Cosh | 23.57 | 0.69 |
| MSE | 26.83 | 0.72 |
| MAE | 27.28 | 0.75 |
| **MS-SSIM+MSE** | **28.54** | **0.83** |

*A. Loss function*

The SSIM [15] (structural similarity index method) and PSNR [16] (peak signal-to-noise ratio) are simple and general evaluation standards of image quality models. The SSIM algorithm respectively evaluates the similarity of the input and composite image from the three aspects of brightness, contrast, and structure. The range of structural similarity is between 0 and 1, and the larger the value, the more similar two images are. The PSNR is often used to test the quality of signal reconstruction in image compression and other fields, and the value does not exceed 40 dB; the higher the score, the better the image quality.

To evaluate the impacts of different loss functions on the generated images, the reconstructed qualities of 100 test images were calculated, and the reconstructed images were respectively tested with the PSNR and SSIM. Table 2 presents the test results of the image quality after reconstruction with different loss functions and perception distances. It can be concluded from the table that when MS-SSIM+MSE and perception distance were used together as the loss function, the PSNR and SSIM of the generated image reflected the best results.

*B. Universal verification*

To verify the performance of the network on other datasets, all vehicle images were cropped to 512×512 pixels, and divided into six categories: white, black, blue, left, right, and front. First, the change directions when the cars faced left, right, and to the front were obtained. As shown in Figure 7(a), when the image of the car facing the front was input, the composite image could be arbitrarily switched between different orientations. Although the quality of the generated images requires improvement, it is evident from the results that the network learned to change the direction of the car. Additionally, to verify the effectiveness of continuous learning, the car color was used as an extended attribute to synthesize the output images. As can be seen from Figure 7(b), regardless of whether the input image color was black or white, the output image could continuously change to the target color blue.

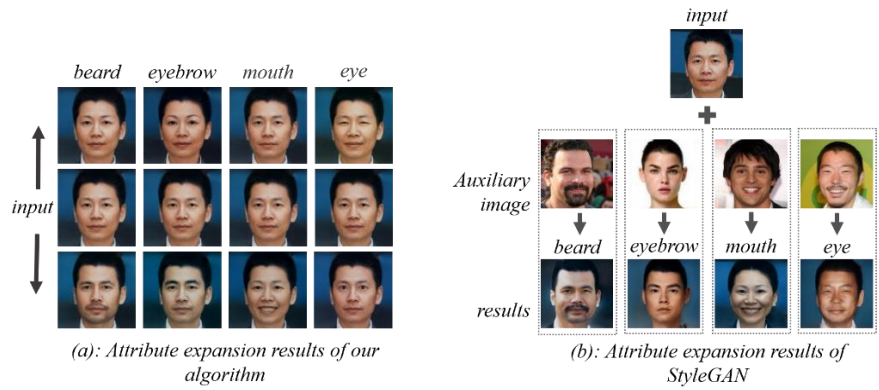

(a): Attribute expansion results of our algorithm

(b): Attribute expansion results of StyleGAN

Fig. 5. The result of the attribute extension

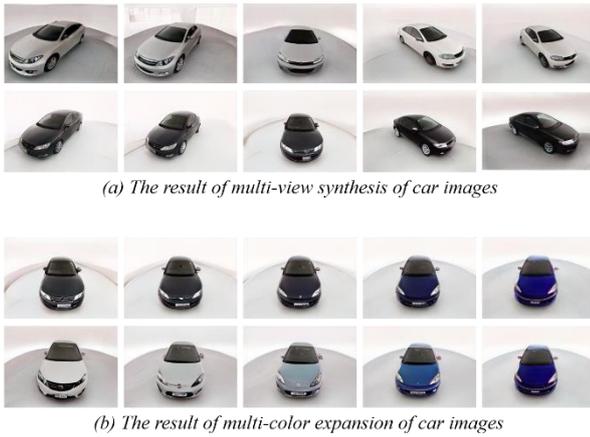

(a) The result of multi-view synthesis of car images

(b) The result of multi-color expansion of car images

Fig. 6. Vehicle attribute manipulation results.

## V. CONCLUSION

In the method proposed in this paper, the feature vector of the input image is first extracted, the change directions of different attributes in feature space are explored, and orthogonal weight modification is then used to decouple the attributes. The continuous synthesis of the target attributes is realized by the linear guidance of input features, and the manipulation process is more controllable. In addition, a new attribute extension module was proposed, which can separate the interference of basic attributes by decomposing new attributes and obtain independent changes of new attributes without retraining the generated network. Via experiments, it was found that the proposed algorithm generate high quality of the synthesized images, the manipulation of attributes is more in line with human sensory logic, and the manipulation of new attributes is more flexible.

## ACKNOWLEDGMENT

This work was supported by the National Nature Science Foundation of China, grant no. 61901436.

## REFERENCES

[1] G. Zeng, Y. Chen, B. Cui, and S. Yu, "Continuous learning of context-dependent processing in neural networks," ArXiv, vol. abs/1810.01256, 2019.

[2] J. N. Kirkpatrick, R. Pascanu et al. "Overcoming catastrophic forgetting in neural networks," Proceedings of the National Academy of Sciences of the United States of America, vol. 114 13, pp. 3521–3526, 2017.

[3] I. J. Goodfellow, Y. Bengio et al. "Generative adversarial nets," in NIPS, 2014.

[4] T. Karras, T. Aila et al, "Progressive growing of gans for improved quality, stability, and variation," ArXiv, vol. abs/1710.10196, 2017.

[5] R. Abdal, Y. Qin, and P. Wonka, "Image2stylegan: How to embed images into the stylegan latent space?" 2019 IEEE/CVF International Conference on Computer Vision (ICCV), pp. 4431–4440, 2019.

[6] T. Karras, S. Laine et al, "Analyzing and improving the image quality of stylegan," ArXiv, vol. abs/1912.04958, 2019.

[7] M. Mirza and S. Osindero, "Conditional generative adversarial nets," ArXiv, vol. abs/1411.1784, 2014.

[8] P. Isola, J.-Y. Zhu, T. Zhou, and A. A. Efros, "Image-to-image translation with conditional adversarial networks," 2017 IEEE Conference on Computer Vision and Pattern Recognition (CVPR), pp. 5967–5976, 2016.

[9] Y. Choi, M.-J. Choi et al, "Stargan: Unified generative adversarial networks for multi-domain image-to-image translation," 2018 IEEE/CVF Conference on Computer Vision and Pattern Recognition, pp. 8789–8797, 2017.

[10] C. H. Lin, E. Yumer et al. "St-gan: Spatial transformer generative adversarial networks for image compositing," 2018 IEEE/CVF Conference on Computer Vision and Pattern Recognition, pp. 9455–9464, 2018.

[11] M.-Y. Liu, T. Breuel, and J. Kautz, "Unsupervised image-to-image translation networks," in NIPS, 2017.

[12] H. Qassim, A. Verma, and D. Feinzimer, "Compressed residual-vgg16 cnn model for big data places image recognition," 2018 IEEE 8th Annual Computing and Communication Workshop and Conference (CCWC), pp.169–175, 2018.

[13] Z. Liu, P. Luo, X. Wang, and X. Tang, "Deep learning face attributes in the wild," in Proceedings of International Conference on Computer Vision (ICCV), December 2015.

[14] A. Mollahosseini, B. Hasani, and M. H. Mahoor, "Affectnet: A database for facial expression, valence, and arousal computing in the wild," IEEE Transactions on Affective Computing, vol. 10, pp. 18–31, 2019.

[15] S. Heng, "Survey of super-resolution image reconstruction methods," 2013.

[16] H. R. Sheikh, A. C. Bovik, and G. de Veciana, "An information fidelity criterion for image quality assessment using natural scene statistics," IEEE Transactions on Image Processing, vol. 14, pp. 2117–2128, 2005.

[17] Z. He, W. Zuo, M. Kan, S. Shan, and X. Chen, "Attgan: Facial attribute editing by only changing what you want," IEEE Transactions on Image Processing, vol. 28, pp. 5464–5478, 2017.